\documentclass[runningheads]{llncs}
\usepackage{amsmath,amsfonts,bm}

\DeclareMathOperator*{\cL}{\mathcal{L}}

\def\1{\bm{1}}

\def\rx{{\textnormal{x}}}
\def\ry{{\textnormal{y}}}

\def\vtheta{{\bm{\theta}}}

\def\vq{{\bm{q}}}

\def\vx{{\bm{x}}}
\def\vy{{\bm{y}}}

\DeclareMathAlphabet{\mathsfit}{\encodingdefault}{\sfdefault}{m}{sl}
\SetMathAlphabet{\mathsfit}{bold}{\encodingdefault}{\sfdefault}{bx}{n}

\newcommand{\softmax}{\mathrm{softmax}}

\usepackage{amssymb}
\usepackage{dsfont}
\usepackage{graphicx}
\usepackage{mathtools}
\usepackage{physics}
\usepackage{textcomp}
\usepackage{amsmath}
\usepackage{algorithm}
\usepackage[noend]{algpseudocode}
\usepackage[dvipsnames]{xcolor}
\usepackage{amsopn}
\usepackage{tabularx,booktabs,subcaption}
\usepackage{hyperref}
\usepackage{array}
\usepackage{upgreek}
\usepackage{footnote}

\usepackage{xr}
\newcolumntype{Y}{>{\centering\arraybackslash}X}
\newcolumntype{C}[1]{>{\centering\let\newline\\\arraybackslash\hspace{0pt}}m{#1}}

\newcommand{\de}{{\beta}}

\newcommand{\uzl}{UHL and MUV}
\begin{document}
\title{\texorpdfstring{Distributionally Robust Segmentation \\of Abnormal Fetal Brain 3D MRI}
{Distributionally Robust Segmentation of Abnormal Fetal Brain 3D MRI}}
%
\titlerunning{Abnormal Fetal Brain 3D MRI Segmentation}
%
\author{Lucas Fidon\inst{1}
    \and Michael Aertsen\inst{2}
    \and Nada Mufti\inst{1,3,4}
    \and Thomas Deprest\inst{2}
    \and \\Doaa Emam\inst{4,6}
    \and Fr\'ed\'eric Guffens\inst{2}
    \and Ernst Schwartz\inst{5}
    \and Michael Ebner\inst{1}
    \and \\Daniela Prayer\inst{5}
    \and Gregor Kasprian\inst{5}
    \and Anna L. David\inst{3,4}
    \and Andrew Melbourne\inst{1}
	\and S\'ebastien Ourselin\inst{1}
	\and Jan Deprest\inst{2,3,4}
	\and Georg Langs\inst{5}
	\and Tom Vercauteren\inst{1}}

\authorrunning{Lucas Fidon et al.}

\institute{
School of Biomedical Engineering \& Imaging Sciences, King's College London, UK 
\and
Department of Radiology, University Hospitals Leuven, Belgium
\and
Institute for Women's Health, University College London, UK
\and
Department of Obstetrics and Gynaecology, University Hospitals Leuven, Belgium
\and
Department of Biomedical Imaging and Image-guided Therapy Medical University of Vienna, Austria
\and
Department of Gynecology and Obstetrics, University Hospitals Tanta, Egypt
}
\maketitle              
\begin{abstract}
The performance of deep neural networks typically increases with the number of training images.
However, not all images have the same importance towards improved performance and robustness.
In fetal brain MRI, abnormalities exacerbate the variability of the developing brain anatomy compared to non-pathological cases.
A small number of abnormal cases, as is typically available in clinical datasets used for training,
are unlikely to fairly represent the rich variability of abnormal developing brains.
This leads machine learning systems trained by maximizing the average performance to be biased toward non-pathological cases.
This problem was recently referred to as hidden stratification.
%
To be suited for clinical use, automatic segmentation methods need to reliably achieve high-quality segmentation outcomes also for pathological cases.
In this paper, we show that the state-of-the-art deep learning pipeline nnU-Net has difficulties to generalize to unseen abnormal cases. 
To mitigate this problem, we propose to train a deep neural network to minimize a percentile of the 
distribution of per-volume loss over the dataset.
%
We show that this can be achieved by using Distributionally Robust Optimization (DRO).
DRO automatically reweights the training samples with lower performance, encouraging nnU-Net to perform more consistently on all cases.
%
We validated our approach using a dataset of 368 fetal brain T2w MRIs, including 124 MRIs of open spina bifida cases and 51 MRIs of cases with other severe abnormalities of brain development.
\end{abstract}
\begin{figure}[tb!]
    \centering
    \includegraphics[width=\textwidth]{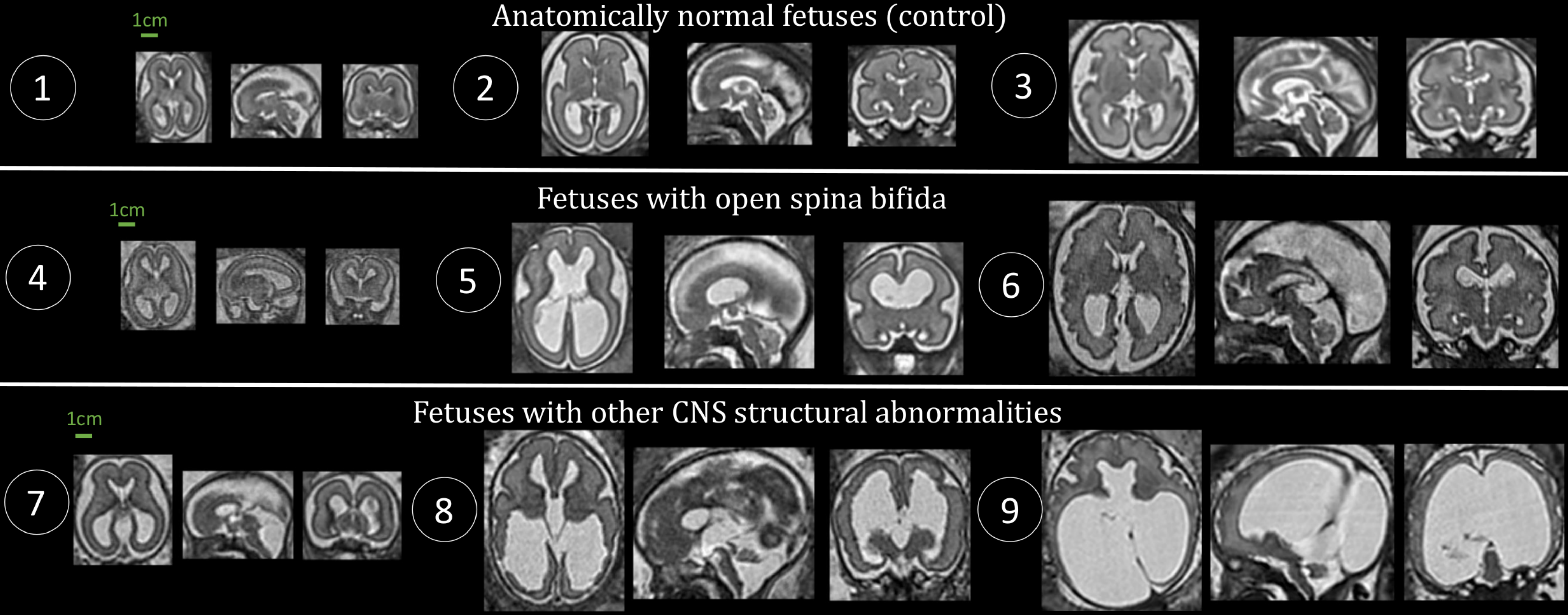}
    \caption{Illustration of the anatomical variability in fetal brain across gestational ages and diagnostics.
    1: Control (22 weeks);  
    2: Control (26 weeks);  
    3: Control (29 weeks);  
    4: Spina bifida (19 weeks);
    5: Spina bifida (26 weeks);
    6: Spina bifida (32 weeks);
    7: Dandy-walker malformation with corpus callosum abnormality (23 weeks);
    8: Dandy-walker malformation with ventriculomegaly and periventricular nodular heterotopia (27 weeks);
    9: Aqueductal stenosis (34 weeks).
    }
    \label{fig:anatomy_variability}
\end{figure}

\section{Introduction}

The segmentation of fetal brain tissues in MRI is essential for 
the study of abnormal fetal brain developments~\cite{benkarim2017toward}.
Fetal brain structures segmentation could also support the evaluation and prediction of surgery outcome for open spina bifida~\cite{aertsen2019reliability,danzer2020fetal,mufti2021cortical,sacco2019fetal,zarutskie2019prenatal}.
Accurate and automatic methods for fetal brain segmentation are necessary as manual segmentation is very time-consuming and suffers from high inter- and intra-rater variability.
Recently, deep neural network-based methods for fetal brain T2w MRI segmentation have been proposed~\cite{fetit2020deep,fidon2021label,khalili2019automatic,payette2021automatic,payette2019longitudinal}.
On average,
deep learning currently achieves state-of-the-art segmentation performance.
However, those studies do not evaluate specifically the generalization and robustness 
properties
when applied to fetuses with a pathological central nervous system.

Datasets used to train deep neural networks typically contain some underrepresented subsets of cases.
These cases are not specifically dealt with by the training algorithms currently used for deep neural networks.
This problem has been referred to as hidden stratification~\cite{oakden2020hidden}.
Hidden stratification has been shown to lead to deep learning models with good average performance but poor performance on some clinically relevant subsets of the population~\cite{oakden2020hidden}.
%
While uncovering the issue,
the study of~\cite{oakden2020hidden}, which is limited to classification, does not study the cause or propose a method to mitigate this problem.
Cases with abnormal fetal brain development are likely to suffer from hidden stratification effects for two reasons:
1) The presence of abnormalities exacerbates the anatomical variability of the fetal brain between 18 weeks and 38 weeks of gestation, as illustrated in Fig.~\ref{fig:anatomy_variability}; and
2) The prevalence of those diseases is typically below 1/1000 births~\cite{aertsen2019reliability}.
%

In this work, we study the problem of hidden stratification in fetal brain MRI segmentation using deep learning.
We claim that the methodology currently used to train deep neural networks, that is maximizing the average performance across the training volumes, is at the root of the hidden stratification problem.
Instead of the average empirical risk, training safe and robust deep learning models requires an asymmetric measure of risk that gives higher weights to the cases for which the algorithm fails (hard examples).
Percentiles, also known as value-at-risk, is such a measure of risk that has even been adopted in industry regulations~\cite{Holton2003ValueAR}.
%
Given a per-volume fetal brain MRI segmentation metric such as the Dice score and an algorithm,
the percentile at $5\%$ is the value of the score below which $5\%$ of the cases fall, i.e. perform worse than the percentile.
%
The percentile relates to hidden stratification effects as it informs us of how badly worst-case examples are performing.
%
Our contributions are four-fold.
1) We empirically show that the state-of-the-art deep learning pipeline nnU-Net~\cite{isensee2021nnu} trained by maximizing the average segmentation performance leads to clinically significant failures for fetal brain MRI segmentation.
2) We propose to use percentiles of the Dice score on clinically relevant subpopulations as a measure of hidden stratification effects.
3) We propose to train a deep learning network to minimize a percentile of the per-volume loss function.
4) We propose a relaxation of this optimization problem based on distributionally robust optimization that can be solved efficiently in practice.
We evaluate the proposed methodology for the automatic segmentation of white matter, ventricles, and cerebellum based on fetal brain 3D T2w MRI.
We used a total of $368$ fetal brain 3D MRIs including anatomically normal fetuses, fetuses with open spina bifida, and fetuses with other central nervous system pathologies for gestational ages ranging from $19$ weeks to $39$ weeks.
Our empirical results suggests that the proposed training method based on distributionally robust optimization leads to better percentiles values for abnormal fetuses.
In addition, qualitative results shows that distributionally robust optimization allows to reduce the number of clinically relevant failures of nnU-Net.

\section{Minimization of a Percentile Loss using Distributionally Robust Optimization}
In this section, we study how a deep neural network 
can be trained to minimize percentiles of the loss function using a distributionally robust optimization~(DRO) approach~\cite{fidon2020sgd}.

Standard deep learning training consists in optimizing the parameters $\vtheta$ of a deep neural network $f(\cdot;\vtheta)$ by minimizing the average 
per-example loss $\cL$
\begin{equation}
    \label{eq:erm}
    \min_{\vtheta} \frac{1}{n} \sum_{i=1}^n \cL \left(f(\vx_i;\vtheta), \vy_i\right)
\end{equation}
Within this empirical risk minimization framework, 
$f(\cdot;\vtheta)$ is typically a Convolutional Neural Network (CNN),
$\cL$ is a smooth per-volume loss function, and $\left\{(\vx_i, \vy_i)\right\}_{i=1}^n$ is the training dataset.

In our case, $\vx_i$ are the input 3D fetal brain T2w MRI volumes and $\vy_i$ are the ground-truth manual segmentations.
This approach is the one used to train state-of-the-art deep learning methods for segmentation using stochastic gradient descent~\cite{isensee2021nnu}.
Due to the scarcity and the higher anatomical variability of abnormal cases illustrated in Fig.~\ref{fig:anatomy_variability}, we cannot assume that the set of all possible fetal brain anatomies is sampled uniformly in the training dataset.
However, in \eqref{eq:erm}, all brain volumes are given the same weight equal to $\frac{1}{n}$.

Instead of the average per-volume loss, for robust and safe segmentation,
we argue that it might be more interesting to minimize
the
percentile
$l_{\alpha}$ at $\alpha$ (e.g. 5\%)
of the per-volume loss function.
Formally, this corresponds to the minimization problem
\begin{equation}
    \label{eq:perc}
        \min_{\vtheta,\, l_{\alpha}} \quad l_{\alpha} \qquad
        \textrm{such that}
        \qquad  
        \mathds{P}\left(
            \cL \left(f(\rx;\vtheta), \ry\right) \geq l_{\alpha}
            \right) \leq \alpha
\end{equation}
where $\mathds{P}$ is the empirical distribution defined by the training dataset.
%
In other words,
if $\alpha=0.05$, the optimal $l_{\alpha}^*(\vtheta)$ of \eqref{eq:perc} for a given value set of parameters $\vtheta$ is the value of the loss such that the per-volume loss function is worse than $l_{\alpha}^*(\vtheta)$ $5\%$ of the time.
As a result, training the deep neural network using \eqref{eq:perc} corresponds to minimizing the percentile of the per-volume loss function $l_{\alpha}^*(\vtheta)$.

Unfortunately, the minimization problem \eqref{eq:perc} cannot be solved directly using stochastic gradient descent to train a deep neural network.
We now propose a tractable upper bound for $l_{\alpha}^*(\vtheta)$ and show that it can be solved in practice using distributionally robust optimization~\cite{fidon2020sgd}.

The Chernoff bound~\cite{chernoff1952measure} applied to the per-volume loss function and the empirical training data distribution states that for all $l_{\alpha}$ and $\de>0$
\begin{equation}
    \mathds{P}\left(
            \cL \left(f(\rx;\vtheta), \ry\right) \geq l_{\alpha}
            \right) 
    \leq 
        \frac{\exp\left(-\de l_{\alpha}\right)}{n} 
        \sum_{i=1}^n \exp\left(\de \cL \left(f(\vx_i;\vtheta), \vy_i\right)\right)
\end{equation}
To link this inequality to the minimization problem \eqref{eq:perc}, we set $\de$ such that
\begin{align}
    \alpha &= \frac{\exp\left(-\de \hat{l}_{\alpha}(\vtheta)\right)}{n} 
        \sum_{i=1}^n \exp\left(\de \cL \left(f(\vx_i;\vtheta), \vy_i\right)\right)\\
    \iff 
    \hat{l}_{\alpha}(\vtheta) &= \frac{1}{\de} \log\left(
        \frac{1}{\alpha n}
        \sum_{i=1}^n \exp\left(\de \cL \left(f(\vx_i;\vtheta), \vy_i\right)\right)
    \right)
\end{align}
$\hat{l}_{\alpha}(\vtheta)$ is therefore an upper bound for $l^*_{\alpha}(\vtheta)$, independently to the value of $\vtheta$.
We propose to relax the minimization problem \eqref{eq:perc} by 
\begin{equation}
    \label{eq:expvar}
    \min_{\vtheta} \frac{1}{\de} \log\left(
        \sum_{i=1}^n \exp\left(\de \cL \left(f(\vx_i;\vtheta), \vy_i\right)\right)
    \right)
\end{equation}
where $\de>0$ is a hyperparameter,
and where
the term $\frac{1}{\de} \log\left(\frac{1}{\alpha n}\right)$ was dropped as being independent of $\vtheta$.
%
While in \eqref{eq:expvar}, $\alpha$ does not appear in the optimization problem directly anymore, $\de$ essentially acts as a substitute for $\alpha$.
The higher the value of $\de$, the higher weights the per-volume losses with a high value will have in \eqref{eq:expvar}.

We give a proof in the supplementary material
that \eqref{eq:expvar} is equivalent to solving the distributionally robust optimization problem
\begin{equation}
    \label{eq:dro}
    \min_{\vtheta}\, \max_{\vq \in \Delta_n}
        \left(
        \sum_{i=1}^n q_i \cL\left(f(\vx_i; \vtheta), \vy_i\right)
        - \frac{1}{\de} D_{KL}\left(\vq\, \biggr\Vert\, \frac{1}{n}\mathbf{1}\right)
        \right)
\end{equation}
where a new unknown probabilities vector parameter $\vq$ is introduced, $\frac{1}{n}\mathbf{1}$ denotes the uniform probability vector $\left(\frac{1}{n}, \ldots, \frac{1}{n}\right)$, $D_{KL}$ is the Kullback-Leibler divergence, $\Delta_n$ is the unit $n$-simplex, and $\de > 0$ is a hyperparameter.
$D_{KL}$ measures the dissimilarity between $\vq$ and the uniform probability vector $\frac{1}{n}\mathbf{1}$  that corresponds to assign the same weight $\frac{1}{n}$ to each sample.
Therefore, $\de$ controls how much the samples with a relatively high loss value (hard examples) are weighted.

Recently, hardness weighted sampling~\cite{fidon2020sgd} was introduced as a principled hard example mining method to solve \eqref{eq:dro}.
Here, we proved that it can be used to minimize the proposed relaxed minimization~\eqref{eq:expvar} of the percentile loss problem.

\section{Anatomically Abnormal Fetal Brain T2w MRI Dataset}

\begin{table}[bt]
	\centering
	\caption{
	\textbf{Training and testing dataset details.}
	Other Abn: other brain structural abnormalities.
	There is no overlap of subjects between training and testing.
	}
	\begin{tabularx}{\textwidth}{ *{5}{Y}}
		\toprule
		Train/Test & Origin & Condition & Volumes & Gestational age (in weeks)\\
		\midrule
		Training & Atlas~\cite{gholipour2017normative} & Control & 18 & [21,\,38]\\
		Training & FeTA~\cite{payette2021automatic} & Control & 5 & [22,\,28]\\
		Training & \uzl{} & Control & 116 & [20,\,35]\\
		Training & \uzl{} & Spina Bifida & 28 & [22,\,34]\\
		Training & \uzl{} & Other Abn & 10 & [23,\,35]\\
		\midrule
		Testing & FeTA~\cite{payette2021automatic} & Control & 28 & [20,\,34]\\
		Testing & FeTA~\cite{payette2021automatic} & Spina Bifida & 31 & [22,\,31]\\
		Testing & FeTA~\cite{payette2021automatic} & Other Abn & 16 & [20,\,34]\\
		Testing & \uzl{} & Control & 26 & [26,\,37]\\
		Testing & \uzl{} & Spina Bifida & 65 & [19,\,33]\\
		Testing & \uzl{} & Other Abn & 25 & [21, 40]\\
	\bottomrule
	\end{tabularx}
	\label{tab:data}
\end{table}

In this section, we give details about the fetal brain 3D MRI data, the labelling protocol, and the pre-processing used in our experiments.

\subsubsection{Public Fetal Brain Datasets}
We used the 18 control fetal brain 3D MRI volumes of the spatio-temporal fetal brain atlas\footnote{\url{http://crl.med.harvard.edu/research/fetal_brain_atlas/}}~\cite{gholipour2017normative} for gestational ages ranging from $21$ weeks to $38$ weeks.
We also used $80$ volumes from the publicly available FeTA MICCAI challenge dataset\footnote{DOI: 10.7303/syn25649159}~\cite{payette2021automatic}.
For the $40$ MIAL 3D MRIs, corrections of the segmentations were performed by authors MA, LF, and PD to reduce the variability against the published segmentation guidelines that was released with the FeTA dataset~\cite{payette2021automatic}.
Those corrections were performed as part of our previous work~\cite{fidon2021label} and are publicly available\footnote{DOI: 10.5281/zenodo.5148611}.
Brain masks for the FeTA data were obtained via affine registration using two fetal brain atlases\footnote{DOI: 10.7303/syn25887675}~\cite{fidon2021atlas,gholipour2017normative}.

\subsubsection*{%
Image Acquisition and Preprocessing for the Private Dataset
}
All images in the private dataset were part of routine clinical care and were acquired at \uzl{}
due to congenital malformations seen on ultrasound.

In total, 
93 cases with open spina bifida,
35 cases with other central nervous system pathologies,
and 
142 cases with other malformations, though with normal brain, and referred as controls,
were included.
%
%
The gestational age at MRI ranged from $19$ weeks to $40$ weeks.
We have started to make fetal brain T2w 3D MRIs publicly available\footnote{\url{https://www.cir.meduniwien.ac.at/research/fetal/}}.
For each study, at least three orthogonal T2-weighted HASTE series of the fetal brain were collected on a $1.5$T scanner using an echo time of $133$ms, a repetition time of $1000$ms, with no slice overlap nor gap, pixel size $0.39$mm to $1.48$mm, and slice thickness $2.50$mm to $4.40$mm.
A radiologist attended all the acquisitions for quality control.

The reconstructed fetal brain 3D MRIs were obtained using \texttt{NiftyMIC}~\cite{ebner2020automated} 
a state-of-the-art super resolution and reconstruction algorithm. The volumes were all reconstructed to a resolution of $0.8$ mm isotropic and registered to a fetal brain atlas~\cite{gholipour2017normative}.
%
%
Our pre-processing improves the resolution, and removes motion between neighboring slices and motion artefacts present in the original 2D slices~\cite{ebner2020automated}.
%
%
We used volumetric brain masks to mask the tissues outside the fetal brain.
Those brain masks were obtained using the automatic segmentation method described in~\cite{ebner2020automated,ranzini2021monaifbs}.

\subsubsection{Labelling Protocol.}
The labelling protocol used for white matter, ventricles and cerebellum is the same as in~\cite{payette2021automatic}.
The three tissue types were segmented for our private dataset by a trained obstetrician and medical students under the supervision of a paediatric radiologist specialized in fetal brain anatomy, who quality controlled and corrected all manual segmentations.

\subsubsection{Separation of the Data into Training and Testing}
A summary of the number of fetal brain 3D MRIs used at training and testing for each central nervous system condition can be found in Table~\ref{tab:data}.
The training dataset contains a total of $177$ cases with a majority of $139$ controls and only $38$ abnormal cases which is typical in clinical datasets.
Five controls from the FeTA dataset were added in the training dataset because we found in preliminary experiments that nnU-Net~\cite{isensee2021nnu} fails on most of the FeTA data at testing when it is trained using only data from \uzl{} and the fetal brain atlas~\cite{gholipour2017normative}.
The testing dataset contains $193$ volumes with a majority of abnormal cases which is necessary to cover the anatomical variability of abnormal cases in our evaluation.

\section{Experiments}

\begin{figure}[bt!]
    \centering
    \includegraphics[width=\linewidth]{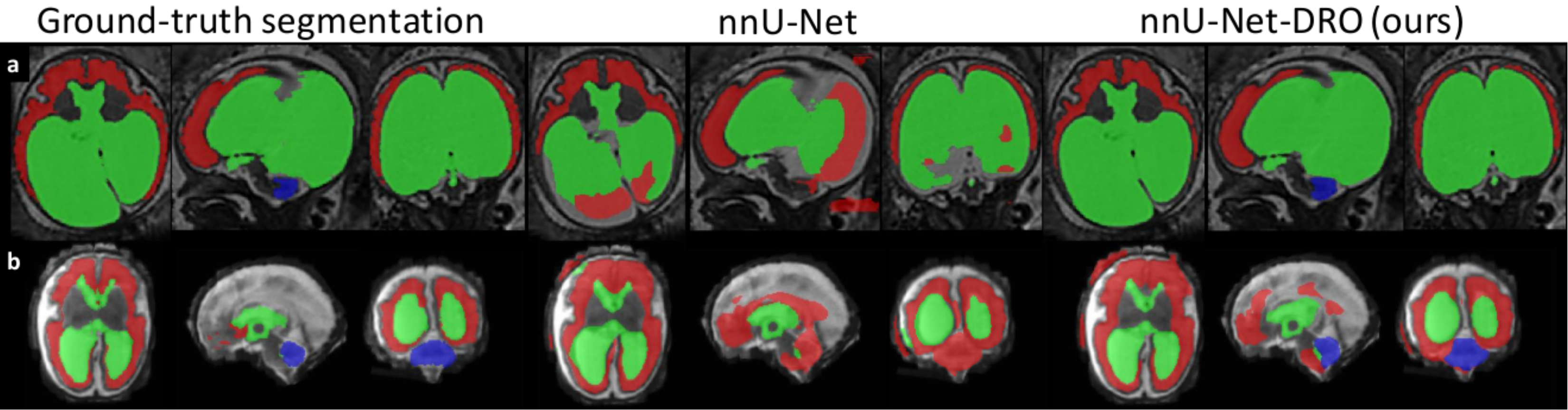}
    \caption{
    \textbf{Qualitative results.}
    a) Fetus with aqueductal stenosis (34 weeks).
    b) Fetus with open spina bifida (27 weeks).
    For those two cases, nnU-Net~\cite{isensee2021nnu} misses completly the cerebellum and achieves poor segmentation for the white matter and the ventricles.
    Our nnU-Net-DRO achieves satisfactory segmentation for the cerebellum for the two cases, and for all tissue types for the aqueductal stenosis case.
    %
    }
    \label{fig:qualitative_results}
\end{figure}

\begin{table}[t]
	\centering
	\caption{\textbf{Evaluation of distribution robustness with respect to the pathology (193 3D MRIs).}
	\textcolor{red}{\bf WM}: White matter, 
	\textcolor{ForestGreen}{\bf Vent}: Ventricles, 
	\textcolor{blue}{\bf Cer}: Cerebellum.
	$\textbf{p}_{X}$: $X^{\textrm{th}}$ percentile of the Dice score distribution in percentage.
    Best values are in bold.
	}
	\begin{tabularx}{\textwidth}{c c c *{6}{Y}}
		\toprule
        \multicolumn{1}{c}{} & \multicolumn{1}{c}{} & \multicolumn{1}{c}{}
        & \multicolumn{6}{c}{Dice Score ($\%$)} \\
        \cmidrule(lr){4-9} 
		\multicolumn{1}{c}{\bf Method} & \multicolumn{1}{c}{\bf CNS} & \multicolumn{1}{c}{\bf ROI} & 
		Mean & Std & $\textbf{p}_{50}$ & $\textbf{p}_{25}$ & $\textbf{p}_{10}$ & $\textbf{p}_5$ \\ 
		\midrule
		(baseline) & \textbf{Controls} &
		        \textcolor{red}{\bf WM} & 
		        $\bf93.9$ & \bf 2.9 & \bf 94.1 & \bf 91.5 & \bf 90.6 & \bf 89.3 \\
		nnU-Net & (54 cases) & \textcolor{ForestGreen}{\bf Vent} &
		        87.8 & 6.8 & 89.7 & 82.1 & 78.1 & \bf 76.8 \\
		        & & \textcolor{blue}{\bf Cer} &
		        \bf 94.5 & 3.2 & \bf 94.6 & 92.4 & \bf 90.7 & \bf 89.8 \\
	        \cmidrule(lr){2-9}
	        & \textbf{Spina Bifida} &
		        \textcolor{red}{\bf WM} & 
		        89.9 & 7.9 & 92.5 & 89.1 & 79.9 & 73.4 \\
		        & (98 cases)  & \textcolor{ForestGreen}{\bf Vent} &
		        90.6 & 10.6 & 93.0 & 88.6 & 84.8 & 80.7 \\
		        & & \textcolor{blue}{\bf Cer} &
		        78.2 & 28.7 & \bf 89.8 & \bf 84.2 & 13.9 & \bf 0.0 \\
		    \cmidrule(lr){2-9}
	        & \textbf{Other Abn.} &
		        \textcolor{red}{\bf WM} & 
		        \bf 90.3 & 9.8 & \bf 92.7 & \bf 89.7 & \bf 82.7 & 70.1 \\
		        & (41 cases) & \textcolor{ForestGreen}{\bf Vent} &
		        87.1 & 7.3 & 87.1 & 82.5 & 77.7 & 75.2 \\
		        & & \textcolor{blue}{\bf Cer} &
		        89.7 & 14.7 & \bf 92.8 & 89.4 & 85.1 & 81.6 \\
	\cmidrule(lr){1-9}
        (ours) & 
		    \textbf{Controls} &
		        \textcolor{red}{\bf WM} & 
		        93.8 & 3.0 & 93.9 & 91.2 & 90.1 & 89.2 \\
		nnU-Net-DRO & (54 cases) & \textcolor{ForestGreen}{\bf Vent} &
		        \bf 87.9 & \bf 6.7 & \bf 89.9 & \bf 82.6 & \bf 78.3 & 76.7 \\
		        & & \textcolor{blue}{\bf Cer} &
		        94.4 & \bf 3.1 & \bf 94.6 & \bf 92.6 & \bf 90.7 & 89.5 \\
	        \cmidrule(lr){2-9}
	        & \textbf{Spina Bifida} &
		        \textcolor{red}{\bf WM} & 
		        $\bf 90.3$ & \bf 7.5 & \bf 92.9 & \bf 89.2 & \bf 81.5 & \bf 73.7\\
		        & (98 cases)  & \textcolor{ForestGreen}{\bf Vent} &
		        $\bf 90.9$ & \bf 10.3 & \bf 93.2 & \bf 89.2 & \bf 85.1 & \bf 81.7 \\
		        & & \textcolor{blue}{\bf Cer} &
		        $\bf 79.7$ & \bf 27.6 & 89.7 & 84.1 & \bf 40.4 & \bf 0.0 \\
		    \cmidrule(lr){2-9}
	        & \textbf{Other Abn.} &
		        \textcolor{red}{\bf WM} & 
		        \bf 90.3 & \bf 9.5 & 92.5 & 89.6 & 82.5 & \bf 72.0 \\
		        & (41 cases) & \textcolor{ForestGreen}{\bf Vent} &
		        \bf 87.5 & \bf 7.1 & \bf 87.5 & \bf 82.7 & \bf 80.4 & \bf 76.7 \\
		        & & \textcolor{blue}{\bf Cer} &
		        \bf 90.6 & \bf 10.5 & \bf 92.8 & \bf 89.8 & \bf 85.5 & \bf 82.9\\
	\bottomrule
	\end{tabularx}
	\label{tab:models_results}
\end{table}

\subsubsection*{Common Deep Learning Pipeline.}
We used nnU-Net~\cite{isensee2021nnu}, a generic deep learning pipeline for medical image segmentation, that has been shown to outperform other deep learning pipelines on 23 public datasets without the need to tune the loss function or the deep neural network architecture.
Specifically, we used nnU-Net version 2 in 3D-full-resolution mode which is the recommended mode for isotropic 3D MRI data.
nnU-Net automatically splits the training data into 5 folds $80\%$ training/$20\%$ validation used to train 5 networks for each method.
The predicted class probability maps of the 5 models are averaged at inference to improve robustness~\cite{isensee2021nnu}.
We used NVIDIA Tesla V100 GPUs with 16GB of memory.
Training each network took from 4 to 6 days.

\subsubsection{Specificities of Each Method.}
The baseline consists in using nnU-Net~\cite{isensee2021nnu} without any modification.
Our method, nnU-Net-DRO, also uses nnU-Net. 
The only difference is that we changed the sampling strategy to use the hardness weighted sampler for DRO~\cite{fidon2020sgd}.
We used the default hyper-parameter values for the hardness weighted sampler, i.e. $\beta=100$ with importance sampling and clipping values $w_{min}=0.1$ and $w_{max}=10$ as described in~\cite{fidon2020sgd}.
No other values were tested.
Our implementation of the nnU-Net-DRO training procedure is publicly available at \url{https://github.com/LucasFidon/HardnessWeightedSampler}.
It provides an implementation of the hardness weighted sampler described in~\cite{fidon2020sgd}.

\subsubsection*{Evaluation Method.}
We evaluate the quality of the automatic fetal brain MRI segmentations using the Dice score~\cite{dice1945measures,fidon2017generalised}.
We are particularly interested in measuring the statistical risk of the results as a way to evaluate the robustness of the different methods.
To this end, in addition to the mean and standard deviation, we also report the percentiles of the Dice score at $50\%$, $25\%$, $10\%$, and $5\%$.
In Table~\ref{tab:models_results}, we report those quantities for the Dice scores of the three tissue types white matter, ventricular system, and cerebellum.

For each method, nnU-Net is trained 5 times using different train/validation splits and different random initializations.
The 5 same splits, computed randomly, are used for the two methods.
The results in Table~\ref{tab:models_results} are for the ensemble of the 5 3D U-Nets.
Ensembling is known to increase the robustness of deep learning methods for segmentation~\cite{isensee2021nnu}.
It also makes the evaluation less sensitive to the random initialization and to the stochastic optimization.

\subsubsection*{Evaluation of nnU-Net and nnU-Net-DRO.}
Quantitative evaluation of nnU-Net and nnU-Net-DRO for the three different central nervous system conditions control, spina bifida, and other abnormalities can be found in Table~\ref{tab:models_results}.

For spina bifida and other brain abnormalities, the proposed nnU-Net-DRO achieves same or higher mean Dice scores and lower standard deviations than nnU-Net~\cite{isensee2021nnu} for the three tissue types.
For controls, the mean Dice scores and standard deviation of nnU-Net-DRO and nnU-Net differ by less than $0.1$ percentage points (pp) for the three tissue types.

The comparison of the percentiles of the Dice score allows us to compare methods at the tail of the Dice scores distribution where segmentation methods reach their worst-case performance.
For spina bifida, nnU-Net-DRO achieves higher values of percentiles than nnU-Net for the white matter ($+0.6$pp for $\textbf{p}_{10}$), for the ventricular system ($+1.0$pp for $\textbf{p}_{5}$), and for the cerebellum ($+26.5$pp for $\textbf{p}_{10}$).
And for other brain abnormalities, nnU-Net-DRO achieves higher values of percentiles than nnU-Net for the white matter ($+1.9$pp for $\textbf{p}_{5}$), for the ventricular system ($+1.5$pp for $\textbf{p}_{5}$ and $+2.7$pp for $\textbf{p}_{10}$), and for the cerebellum ($+1.3$pp for $\textbf{p}_{5}$).
All the other percentile values differ by less than $0.5$pp of Dice score between the two methods.
This suggests that nnU-Net-DRO achieves better worst case performance than nnU-Net for abnormal cases.

It is worth noting that the Dice scores decrease for the white matter and the cerebellum between controls and spina bifida and abnormal cases.
It was expected due to the higher anatomical variability in pathological cases.
However, the Dice scores for the ventricular system tend to be higher for abnormal cases than for controls.
This can be attributed to the large proportion of pathological cases with enlarged ventricles because the Dice score values tend to be higher for larger region of interests.

As can be seen in the qualitative results of Table~\ref{fig:qualitative_results}, there are cases for which nnU-Net predicts an empty cerebellum segmentation while nnU-Net-DRO achieves satisfactory cerebellum segmentation.
There were no cases for which the converse was true.
%
%
Robust segmentation of the cerebellum for spina bifida is particularly relevant for the evaluation of fetal brain surgery for open spina bifida~\cite{aertsen2019reliability,danzer2020fetal,sacco2019fetal}.
Additional qualitative results in the supplementary material
illustrates 5 other cases for which nnU-Net-DRO outperforms nnU-Net.

\section{Conclusion}
The high anatomical variability of the developing fetal brain across gestational ages and pathologies hampers the robustness of deep neural networks trained by maximizing the average per-volume performance.
Specifically, it limits the generalization of deep neural networks to abnormal cases for which few cases are available during training.
In this paper, we propose to mitigate this problem by training deep neural networks to minimize a percentile of the per-volume performance rather than the average.
To allow to do this in practice,
we propose to train deep neural networks with Distributionally Robust Optimization (DRO) and we show that the DRO objective is a relaxation of the per-volume loss percentile.
We have validated the proposed training method on a multi-centric dataset of $368$ fetal brain T2w 3D MRIs with various diagnostics.
nnU-Net trained with DRO achieved improved segmentation results for pathological cases as compared to the unmodified nnU-Net, while achieving similar segmentation performance for the neurotypical cases.
Our results suggest that nnU-Net trained with DRO is more robust to anatomical variabilities than the original nnU-Net.

\subsubsection*{Acknowledgments}
This project has received funding from the European Union's Horizon 2020 research and innovation program under the Marie Sk{\l}odowska-Curie grant agreement TRABIT No 765148.
This work was supported by core and project funding from the
Wellcome [203148/Z/16/Z; 203145Z/16/Z; WT101957], and EPSRC [NS/A000049/1; NS/A000050/1; NS/A000027/1].
TV is supported by a Medtronic / RAEng Research Chair [RCSRF1819\textbackslash7\textbackslash34].

%
%
%
\bibliographystyle{splncs04.bst}
\bibliography{main.bib}

\newpage
\begin{center}
\textbf{\Large --- Supplemental Document ---\\Distributionally Robust Segmentation\\
of Abnormal Fetal Brain 3D MRI}\\[.6cm]
Lucas Fidon,$^{1}$ Michael Aertsen,$^{2}$ Nada Mufti,$^{1,3,4}$ Thomas Deprest,$^{2}$\\Doaa Emam,$^{4,6}$ Fr\'ed\'eric Guffens,$^{2}$ Ernst Schwartz,$^{5}$ Michael Ebner,$^{1}$\\ Daniela Prayer,$^{5}$ Gregor Kasprian,$^{5}$ Anna L. David,$^{3,4}$ Andrew Melbourne,$^{1}$\\ S\'ebastien Ourselin,$^{1}$ Jan Deprest,$^{2,3,4}$ Georg Langs,$^{5}$ and Tom Vercauteren$^{1}$\\[.3cm]
  \small ${}^1$School of Biomedical Engineering \& Imaging Sciences, King’s College London, UK\\
  ${}^2$Department of Radiology, University Hospitals Leuven, Belgium\\
  ${}^3$Institute for Women's Health, University College London, UK\\
  ${}^4$Department of Obstetrics and Gynaecology, University Hospitals Leuven, Belgium\\
  ${}^5$Department of Biomedical Imaging and Image-guided Therapy Medical University of Vienna, Austria\\
  ${}^6$Department of Gynecology and Obstetrics, University Hospitals Tanta, Egypt
\end{center}

\setcounter{figure}{0}
\setcounter{table}{0}
\setcounter{page}{1}
\setcounter{section}{0}
\makeatletter

\section{Proof of the equivalence of equations (6) and (7)}

In the DRO optimization problem of equation (7), the optimal $\vq$ for any $\vtheta$ has the closed-form formula~\cite[see p.4 or Appendix 11.1]{fidon2020sgd}
\begin{equation*}
    \forall \vtheta,\quad
    \vq^*\left(\vtheta\right) = \softmax\left(
        \left(\de \cL\left(h(\vx_i; \vtheta), \vy_i\right)\right)_{i=1}^n
    \right)
\end{equation*}
By injecting this in equation (7), we obtain
\begin{align*}
    \min_{\vtheta}\,& \max_{\vq \in \Delta_n}
        \left(
        \sum_{i=1}^n q_i \cL\left(f(\vx_i; \vtheta), \vy_i\right)
        - \frac{1}{\de} D_{KL}\left(\vq\, \biggr\Vert\, \frac{1}{n}\mathbf{1}\right)
        \right)\\
    \min_{\vtheta}\,&
        \left(
        \sum_{i=1}^n q^*_i(\vtheta) \cL\left(f(\vx_i; \vtheta), \vy_i\right)
        - \frac{1}{\de} 
            \sum_{i=1}^n q^*_i(\vtheta)
                \log\left(\frac{
                    \exp\left(\de \cL\left(f(\vx_i; \vtheta), \vy_i\right)\right)}{
                    \frac{1}{n}\sum_{j=1}^n\exp\left(\de \cL\left(f(\vx_j; \vtheta), \vy_j\right)\right)}\right)
        \right)\\
    \min_{\vtheta}\,&
        \left(
        \sum_{i=1}^n q^*_i(\vtheta) \cL\left(f(\vx_i; \vtheta), \vy_i\right)
        - \sum_{i=1}^n q^*_i(\vtheta)
                \frac{1}{\de} \log\left(\exp\left(\de \cL\left(f(\vx_i; \vtheta), \vy_i\right)\right)\right)
        \right.\\
    & + \frac{1}{\de}
        \left.
            \left(\sum_{i=1}^n q^*_i(\vtheta)\right) \times
            \log\left(
            \frac{1}{n}\sum_{j=1}^n\exp\left(\de \cL\left(f(\vx_j; \vtheta), \vy_j\right)\right)
            \right)
        \right)\\
    %
\end{align*}
Since the first two terms cancel each other and $\sum_{i=1}^n q^*_i(\vtheta)=1$, we obtain
\begin{align*}
    \min_{\vtheta}\,& \frac{1}{\de} \log\left(
            \sum_{j=1}^n\exp\left(\de \cL\left(f(\vx_j; \vtheta), \vy_j\right)\right)
            \right)
            - \frac{1}{\de} \log\left(n\right)
\end{align*}
which is equivalent to the optimization problem (6) because the term $\frac{1}{\de} \log\left(n\right)$ above and the term $\frac{1}{\de} \log\left(\alpha n\right)$ in (6) are independent of $\vtheta$ $\blacksquare$

\clearpage

\section{Additional qualitative results}
\begin{figure}[h]
    \centering
    \includegraphics[width=\textwidth]{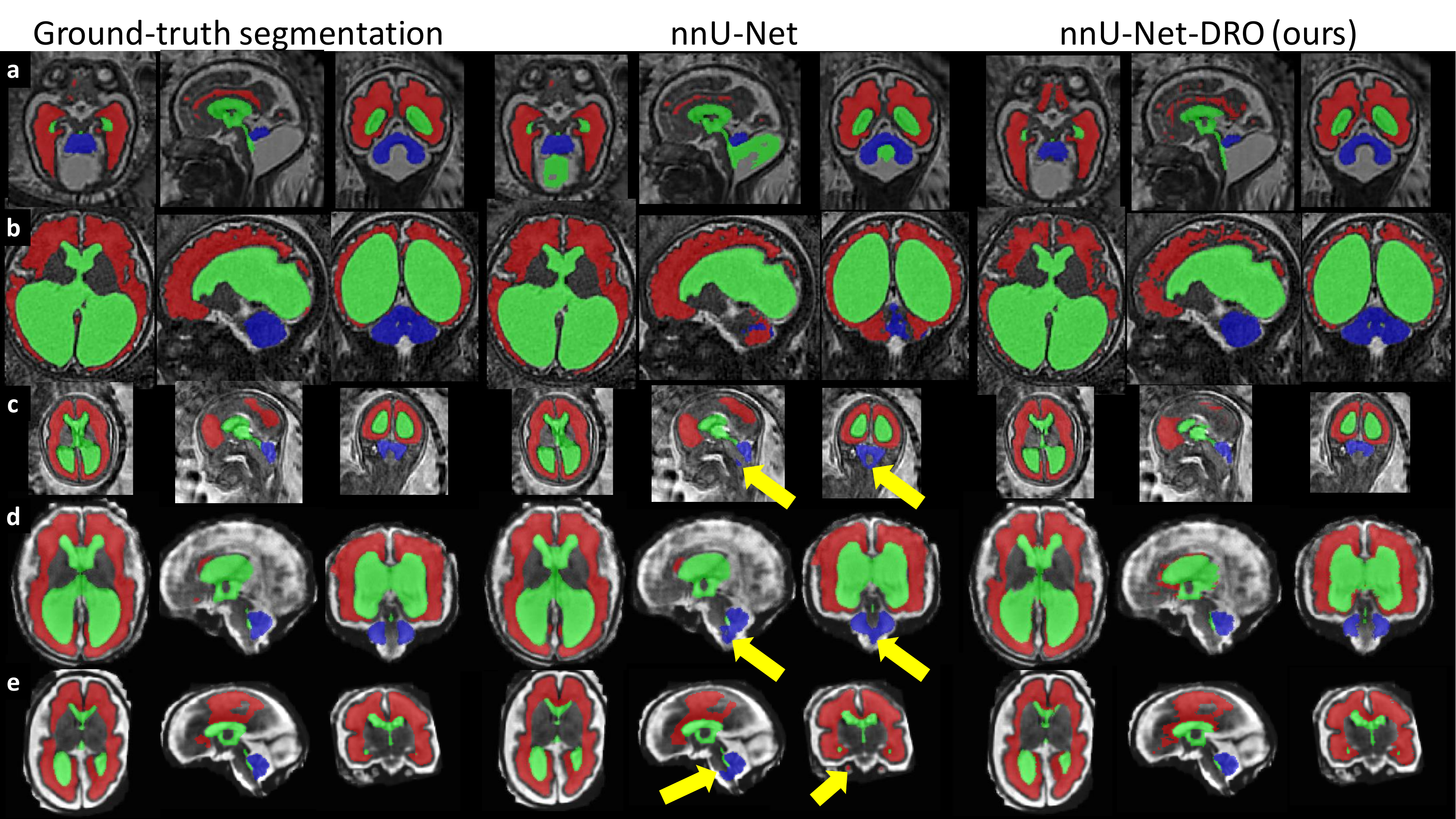}
    \caption{\textbf{Additional qualitative results.}
    a) Fetus with Blake's pouch cyst (29 weeks).
    b) Fetus with tuberous sclerosis complex (34 weeks).
    c) Fetus with open spina bifida (22 weeks).
    d) Fetus with open spina bifida (31 weeks).
    e) Fetus with open spina bifida (28 weeks).
    For case a), a large part of the Blake's pouch cyst is wrongly included in the ventricular system segmentation by nnU-Net~\cite{isensee2021nnu}. This is not the case for the proposed nnU-Net-DRO.
    For case b), nnU-Net fails to segment the cerebellum correctly and a large part of the cerebellum is segmented as part of the white matter. In contrast, our nnU-Net-DRO correctly segment cerebellum and white matter for this case.
    For cases c) d) and e), we have added yellow arrows pointing to large parts of the brainstem that nnU-Net wrongly included in the cerebellum segmentation. 
    nnU-Net-DRO does not make this mistake.
    We emphasise that the segmentation of the cerebellum for open spina bifida is essential for studying and evaluating the effect of surgery in-utero.
    }
    \label{fig:res_qualitative}
\end{figure}

\end{document}